\documentclass[10pt,twocolumn,a4paper]{article}

\usepackage[utf8]{inputenc}
\usepackage[T1]{fontenc}
\usepackage{mathptmx}
\usepackage{amsmath,amssymb}
\usepackage{graphicx}
\usepackage{booktabs}
\usepackage{multirow}
\usepackage{enumitem}
\usepackage[margin=1.8cm,columnsep=0.6cm]{geometry}
\usepackage[hidelinks]{hyperref}
\usepackage{url}
\usepackage{natbib}
\usepackage{lineno}

\title{\textbf{Memory for Autonomous LLM Agents:\\Mechanisms, Evaluation, and Emerging Frontiers}}

\author{
  Pengfei Du\textsuperscript{1}\\[3pt]
  \small\textsuperscript{1} Hong Kong Research Institute of Technology, HongKong, China \\[1pt]
  \small E-mail: lldpf1234@gmail.com
  }

\date{}

\begin{document}
\maketitle

\begin{abstract}
\noindent
Large language model (LLM) agents increasingly operate in settings where a single context window is far too small to capture what has happened, what was learned, and what should not be repeated.
Memory---the ability to persist, organize, and selectively recall information across interactions---is what turns a stateless text generator into a genuinely adaptive agent.
This survey offers a structured account of how memory is designed, implemented, and evaluated in modern LLM-based agents, covering work from 2022 through early 2026.
We formalize agent memory as a \emph{write--manage--read} loop tightly coupled with perception and action, then introduce a three-dimensional taxonomy spanning temporal scope, representational substrate, and control policy.
Five mechanism families are examined in depth: context-resident compression, retrieval-augmented stores, reflective self-improvement, hierarchical virtual context, and policy-learned management.
On the evaluation side, we trace the shift from static recall benchmarks to multi-session agentic tests that interleave memory with decision-making, analyzing four recent benchmarks that expose stubborn gaps in current systems.
We also survey applications where memory is the differentiating factor---personal assistants, coding agents, open-world games, scientific reasoning, and multi-agent teamwork---and address the engineering realities of write-path filtering, contradiction handling, latency budgets, and privacy governance.
The paper closes with open challenges: continual consolidation, causally grounded retrieval, trustworthy reflection, learned forgetting, and multimodal embodied memory.
\end{abstract}


\section{Introduction}
\label{sec:intro}

Scaling large language models has unlocked a new class of autonomous software agents---systems that perceive environments, reason about goals, wield tools, and take action over extended time horizons~\citep{brown2020gpt3,achiam2023gpt4,touvron2023llama}.
What separates these agents from a vanilla chatbot is not merely bigger models; it is the expectation that they \emph{learn from experience}.
A coding assistant should remember that a particular API is flaky, a game-playing agent should recall which crafting recipes it already mastered, and a personal scheduler should never ask a user's birthday twice.
All of this demands memory.

\subsection{What goes wrong without it}
\label{sec:why_memory}

Picture a debugging assistant that works on a large codebase across a week of sessions.
Without memory, every Monday morning it rediscovers the directory layout, re-reads the same README, and---worst of all---retries the exact fix that crashed the build on Friday.
Equip the same agent with even a modest memory module and the dynamic shifts: it arrives already knowing the hotspots, skips the dead ends, and gradually distills project-specific heuristics.

This is not a marginal improvement; it is a qualitative change.
Memory transforms a stateless LLM into a \emph{self-evolving} agent~\citep{zhang2024surveymemory} that can (i)~accumulate factual knowledge and user preferences, (ii)~develop behavioral patterns grounded in prior experience, (iii)~avoid repeating costly mistakes, and (iv)~continuously improve through interaction.

\subsection{A brief history of neural memory}
\label{sec:history}

The ambition to give neural networks external storage dates back over a decade.
Memory Networks~\citep{weston2015memnn} and their end-to-end variant~\citep{sukhbaatar2015memn2n} introduced differentiable read-write access to external slots for question answering.
Neural Turing Machines~\citep{graves2014ntm} and Differentiable Neural Computers~\citep{graves2016dnc} pushed further, supporting both content-based and location-based addressing over a memory matrix.
Memorizing Transformers~\citep{wu2022memorizing} and Recurrent Memory Transformers~\citep{bulatov2022recurrentmem} later integrated explicit memory layers directly into the Transformer backbone.

A parallel thread focused on retrieval.
RAG~\citep{lewis2020rag} married a pre-trained generator with a dense document retriever, and RETRO~\citep{borgeaud2022retro} showed that pulling from a \emph{trillion}-token corpus at inference time could match much larger models at a fraction of the parameter count.
These systems proved a crucial point: external knowledge stores can be queried dynamically during generation without retraining.

The leap from retrieval-augmented \emph{models} to memory-augmented \emph{agents} happened quickly.
ReAct~\citep{yao2022react} interleaved reasoning traces with environment actions, producing an interpretable trajectory that doubles as short-horizon memory.
Reflexion~\citep{shinn2023reflexion} took this further by storing verbal self-critiques after task attempts---essentially giving the agent a post-mortem journal.
Then came the Generative Agents paper~\citep{park2023generative}, whose simulated town of 25 characters demonstrated that a simple observation--reflection--planning loop could produce months of coherent social behavior.
Since 2023, the design space has exploded: hierarchical virtual memory inspired by operating systems~\citep{packer2024memgpt}, ever-growing skill libraries in Minecraft~\citep{wang2023voyager}, SQL databases as symbolic memory~\citep{hu2023chatdb}, and---most recently---end-to-end learned memory management via reinforcement learning~\citep{yu2026agentic}.

\subsection{Why another survey?}
\label{sec:motivation}

Several broad agent surveys already exist~\citep{xi2023rise,wang2024survey}, and Zhang et al.~\citep{zhang2024surveymemory} published a memory-focused review in 2024.
However, the landscape has shifted considerably since then.
A wave of 2025--2026 contributions---Agentic Memory~\citep{yu2026agentic}, MemBench~\citep{tan2025membench}, MemoryAgentBench~\citep{hu2025memoryagentbench}, MemoryArena~\citep{he2026memoryarena}---has introduced learned memory control, richer evaluation dimensions, and agentic benchmarks that tightly couple memory with action.

This survey zooms in on the \emph{memory module} and asks three questions:
\begin{enumerate}[label=\textbf{RQ\arabic*},leftmargin=2.5em]
    \item How should memory in LLM agents be decomposed and formalized?
    \item What mechanisms exist, and what trade-offs do they impose?
    \item How should memory be evaluated when the ultimate test is downstream agent performance?
\end{enumerate}

\noindent\textbf{Contributions.}
We formalize agent memory as a write--manage--read loop within a POMDP-style agent cycle (Section~\ref{sec:formulation}), propose a three-dimensional taxonomy that unifies disparate designs (Section~\ref{sec:taxonomy}), provide deep mechanism reviews with concrete system comparisons (Section~\ref{sec:mechanisms}), survey benchmarks alongside a practical metric stack (Section~\ref{sec:evaluation}), map applications where memory is the differentiating factor (Section~\ref{sec:applications}), discuss engineering realities and architecture patterns (Section~\ref{sec:systems}), position relative to prior surveys (Section~\ref{sec:related}), and chart open research directions (Section~\ref{sec:future}).

\section{Problem Formulation and Design Objectives}
\label{sec:formulation}

\subsection{The agent loop, seen through memory}
\label{sec:agentloop}

At each discrete step~$t$, an agent receives input $x_t$---a user message, a sensor reading, or a tool return value---and must produce an action~$a_t$.
Between these two events, it consults its accumulated memory.
We write:
\begin{align}
a_t &= \pi_\theta\!\bigl(x_t,\;\mathcal{R}(M_t, x_t),\;g_t\bigr), \label{eq:action} \\
M_{t+1} &= \mathcal{U}\!\bigl(M_t, x_t, a_t, o_t, r_t\bigr), \label{eq:memupdate}
\end{align}
where $\pi_\theta$ is the policy (typically a prompted or partially fine-tuned LLM), $\mathcal{R}$ reads from memory, $\mathcal{U}$ writes to and manages memory, $g_t$ encodes active goals, $o_t$ is environment feedback, and $r_t$ is any reward-like signal.

Two aspects deserve emphasis.
First, $\mathcal{U}$ is \emph{not} a simple append operation.
In a well-designed system it summarizes, deduplicates, scores priority, resolves contradictions, and---when appropriate---deletes.
Second, $\pi_\theta$ and $(\mathcal{R},\mathcal{U})$ form a feedback loop: the agent's decisions determine what gets written, and what is written shapes future decisions.
This recursive dependence is what makes memory both powerful and brittle---one bad write can pollute the store for many steps downstream.

\subsection{Connection to POMDPs}
\label{sec:pomdp}

Cast formally, the setup above is a partially observable Markov decision process.
Memory $M_t$ plays the role of the agent's \emph{belief state}: an internal summary of history that stands in for the unobservable true state of the world.
Classical POMDP solvers update beliefs via Bayesian filtering; LLM agents do something analogous---albeit messier---through natural language compression, vector indexing, or structured storage.

The analogy clarifies an important point: agent memory is not merely a database lookup problem.
It is about maintaining a \emph{sufficient statistic} of the interaction history for good action selection, subject to hard computational and storage budgets.

\subsection{Five design objectives and their tensions}
\label{sec:objectives}

Across the systems we review, memory mechanisms are pulled along five axes:

\begin{itemize}[leftmargin=1.5em]
    \item \textbf{Utility} -- Does memory actually improve task outcomes?
    \item \textbf{Efficiency} -- What is the token, latency, and storage cost per unit of utility gained?
    \item \textbf{Adaptivity} -- Can the system update incrementally from interaction feedback without a full retrain?
    \item \textbf{Faithfulness} -- Is recalled information accurate and current? Stale or hallucinated recall can be worse than no recall at all.
    \item \textbf{Governance} -- Does the system respect privacy, support deletion requests, and comply with organizational policy?
\end{itemize}

These objectives tug in opposite directions.
Maximizing utility tempts you to store everything, which bloats storage and creates governance headaches.
Aggressive compression improves efficiency but silently discards the one rare fact that turns out to be critical three weeks later.
Any real deployment must navigate these trade-offs deliberately, and the ``right'' balance point shifts with the application.
A medical triage agent, where a missed allergy record could be life-threatening, operates under a very different faithfulness--efficiency frontier than a casual recipe recommender.
Understanding these tensions is not merely academic---it directly shapes architectural choices, as we discuss in subsequent sections.

\subsection{Memory as a differentiator: an empirical perspective}
\label{sec:mem_differentiator}

The practical importance of memory design is perhaps best illustrated by ablation results reported across recent systems.
In the Generative Agents experiment~\citep{park2023generative}, removing the reflection component caused agent behavior to degenerate from coherent multi-day planning to repetitive, context-free responses within 48 simulated hours.
Voyager~\citep{wang2023voyager} without its skill library lost 15.3$\times$ in tech-tree milestone speed---the skill library \emph{was} the performance.
And in MemoryArena~\citep{he2026memoryarena}, swapping an active memory agent for a long-context-only baseline dropped task completion from over 80\% to roughly 45\% on interdependent multi-session tasks.

These numbers underscore a recurring theme: the gap between ``has memory'' and ``does not have memory'' is often larger than the gap between different LLM backbones.
Investing in memory architecture can yield returns that rival---or exceed---model scaling.

\section{A Unified Taxonomy of Agent Memory}
\label{sec:taxonomy}

Cognitive scientists have long distinguished multiple memory systems in the human brain~\citep{atkinson1968human,tulving1972episodic,baddeley2000episodic,squire2004memory}.
Agent designers---often unconsciously---mirror that structure.
We organize the space along three orthogonal dimensions.

\subsection{Temporal scope}
\label{sec:temporal}

\textbf{Working memory.}
Whatever fits inside the current context window constitutes the agent's working memory.
Baddeley's central executive plus buffer model~\citep{baddeley2000episodic} maps neatly: the LLM is the executive, the context window is the buffer, and both share the same bottleneck---limited capacity.

\textbf{Episodic memory.}
Records of concrete experiences: individual tool calls, conversation turns, environment observations.
In the Generative Agents world~\citep{park2023generative}, every observation---``Isabella saw Klaus painting in the park at 3pm''---lands in the episodic stream with a timestamp, an importance score, and an embedding for later retrieval.

\textbf{Semantic memory.}
Abstracted, de-contextualized knowledge.
An episodic fact like ``the user corrected the date format on Jan 5, Jan 12, and Feb 1'' may consolidate into the semantic record ``user prefers DD/MM/YYYY.''
This consolidation is rarely automatic; most current systems require explicit prompting or heuristic triggers.

\textbf{Procedural memory.}
Reusable skills and executable plans.
Voyager's skill library~\citep{wang2023voyager} is the clearest example: every verified Minecraft routine is stored as runnable JavaScript, indexed by a natural language description, and composed on the fly for novel tasks.

In practice, most agents blend at least two of these.
The hard question is the \emph{transition policy}: when does an episodic record graduate to semantic status, and when does a semantic fact get instantiated back into working memory for a specific task?

To illustrate the interplay, consider a customer-support agent handling returns.
Each return request constitutes an episodic record.
After processing hundreds of similar requests, the agent might consolidate the pattern into a semantic rule: ``customers who received damaged items within 7 days are eligible for express replacement.''
When a new request arrives, this semantic rule is loaded into working memory alongside the specific episodic details of the current case.
If the agent also has stored scripts for processing returns (procedural memory), the four memory types form a complete reasoning stack: the procedure says \emph{how}, semantic memory says \emph{what the policy is}, episodic memory says \emph{what happened}, and working memory holds the live reasoning context.

This four-layer integration is the aspiration; most current systems implement only two layers well and handle the transitions between layers via crude heuristics.
The consolidation step---where episodes become semantic knowledge---is particularly underserved: it typically requires either explicit developer rules or periodic LLM-driven summarization, both of which are fragile and hard to validate.

\subsection{Representational substrate}
\label{sec:substrate}

How memory is physically stored constrains what the agent can efficiently do with it.

\textbf{Context-resident text}---summaries, scratchpads, chain-of-thought traces~\citep{wei2022chain}---is the simplest substrate.
Fully transparent, zero infrastructure, but ruthlessly capacity-limited.

\textbf{Vector-indexed stores} encode records as dense embeddings and support approximate nearest-neighbor search~\citep{karpukhin2020dense,johnson2019faiss}.
They scale gracefully to millions of records but lose structured relationships: you can ask ``what's most similar?'' but not ``what caused what?''

\textbf{Structured stores}---SQL databases~\citep{hu2023chatdb}, key--value maps, knowledge graphs~\citep{ji2022kgsurvey}---preserve relational structure and support complex queries (``all API failures involving service X in the last 7 days''), at the cost of upfront schema design.

\textbf{Executable repositories}---code libraries, tool definitions, plan templates~\citep{wang2023voyager}---let the agent invoke stored skills directly, sidestepping regeneration and the errors it introduces.

\textbf{Hybrid stores} are the norm in production.
MemGPT~\citep{packer2024memgpt}, for instance, layers a context-window ``main memory'' over a searchable recall database and a vector-indexed archive---each tier with different access patterns and eviction rules.

\subsection{Control policy}
\label{sec:control}

Perhaps the most consequential---and least discussed---dimension is \emph{who decides} what to store, what to retrieve, and what to discard.

\textbf{Heuristic control} hard-codes rules: top-$k$ retrieval, summarize every $n$ turns, expire records older than $d$ days.
Predictable, easy to debug, but blind to context.

\textbf{Prompted self-control} exposes memory operations as tool calls and lets the LLM decide when to invoke them.
MemGPT's \texttt{core\_memory\_append} and \texttt{archival\_memory\_search} are canonical examples~\citep{packer2024memgpt}.
Quality here hinges on the LLM's instruction-following ability and on how well the memory API is documented in the system prompt.

\textbf{Learned control} treats memory operations as policy actions optimized end-to-end.
Agentic Memory~\citep{yu2026agentic} trains store, retrieve, update, summarize, and discard as callable tools via a three-stage RL pipeline with step-wise GRPO.
The payoff is substantial---learned policies discover non-obvious strategies such as preemptive summarization before the context is full---but so is the training cost.

\subsection{Representative systems at a glance}
\label{sec:methods}

Table~\ref{tab:methods} plots key systems and benchmarks on a timeline.

\begin{table*}[!t]
\centering
\caption{Representative memory systems and benchmarks for LLM agents (2020--2026).}
\label{tab:methods}
\small
\begin{tabular}{p{2.8cm}cp{2.8cm}p{7.2cm}}
\toprule
\textbf{System} & \textbf{Year} & \textbf{Memory Category} & \textbf{Distinguishing Feature} \\
\midrule
RAG~\citep{lewis2020rag}
  & 2020 & Non-parametric retrieval
  & First to couple a seq2seq generator with a dense document retriever at NeurIPS 2020. \\[2pt]
RETRO~\citep{borgeaud2022retro}
  & 2022 & Retrieval at scale
  & Chunks retrieved from a 2-trillion-token corpus; 7.5B-parameter model rivals 175B Jurassic-1 on 10/16 benchmarks. \\[2pt]
ReAct~\citep{yao2022react}
  & 2022 & Trajectory traces
  & Reasoning-and-acting traces double as short-horizon working memory; 34\% absolute gain on ALFWorld. \\[2pt]
Reflexion~\citep{shinn2023reflexion}
  & 2023 & Reflective episodic
  & Verbal self-critiques stored as episodic memory; 91\% pass@1 on HumanEval (vs.\ 80\% GPT-4 baseline). \\[2pt]
Generative Agents~\citep{park2023generative}
  & 2023 & Episodic + reflective
  & 25 simulated characters autonomously organize a Valentine's party via observation--reflection--planning cycles. \\[2pt]
Voyager~\citep{wang2023voyager}
  & 2023 & Procedural skill library
  & 3.3$\times$ more unique items and 15.3$\times$ faster tech-tree progression than prior Minecraft agents. \\[2pt]
LongMem~\citep{wang2023longmem}
  & 2023 & Long-form external
  & Frozen backbone + residual side-network; memory bank scales to 65k tokens. \\[2pt]
ChatDB~\citep{hu2023chatdb}
  & 2023 & Structured symbolic
  & SQL databases as agent memory; supports precise INSERT/SELECT queries over interaction records. \\[2pt]
ExpeL~\citep{zhao2024expel}
  & 2024 & Experiential learning
  & Systematically extracts success/failure ``rules of thumb'' from trajectory comparisons. \\[2pt]
MemGPT~\citep{packer2024memgpt}
  & 2024 & Hierarchical virtual
  & OS-inspired paging across main context, recall DB, and archival vector store. \\[2pt]
MemoryBank~\citep{zhong2024memorybank}
  & 2024 & Long-term with forgetting
  & Ebbinghaus-curve decay applied to chatbot memory; published at AAAI 2024. \\[2pt]
LoCoMo~\citep{maharana2024locomo}
  & 2024 & Benchmark
  & Up to 35 sessions, 300+ turns, 9k--16k tokens per conversation; humans still far ahead. \\[2pt]
MemBench~\citep{tan2025membench}
  & 2025 & Benchmark
  & Separates factual vs.\ reflective memory; participation vs.\ observation modes; ACL 2025 Findings. \\[2pt]
MemoryAgentBench~\citep{hu2025memoryagentbench}
  & 2025 & Benchmark
  & Tests four cognitive competencies; no current system masters all four. \\[2pt]
Agentic Memory~\citep{yu2026agentic}
  & 2026 & Unified STM/LTM policy
  & Memory ops trained as RL actions via step-wise GRPO; outperforms all memory-augmented baselines on five benchmarks. \\[2pt]
MemoryArena~\citep{he2026memoryarena}
  & 2026 & Benchmark
  & Multi-session interdependent tasks in four domains; near-saturated LoCoMo models drop to 40--60\% here. \\
\bottomrule
\end{tabular}
\end{table*}

\section{Core Memory Mechanisms}
\label{sec:mechanisms}

We now examine each mechanism family in detail, grounding the discussion in concrete system designs and their empirical trade-offs.

\subsection{Context-resident memory and compression}
\label{sec:context_mem}

The most straightforward way to give an agent memory is to keep relevant information in the prompt.
System messages, recent conversation turns, scratchpad notes---everything the LLM ``sees'' on every call functions as working memory with perfect in-window recall.

The trouble starts when history outgrows the window.
Several compression strategies have emerged:
(i)~\emph{sliding windows} that retain the $n$ most recent turns and drop the rest;
(ii)~\emph{rolling summaries} that periodically condense older history into a shorter precis;
(iii)~\emph{hierarchical summaries} operating at turn, session, and topic granularities;
(iv)~\emph{task-conditioned compression}, where the current query decides which parts of history keep full detail.
The Self-Controlled Memory system~\citep{liang2023encouraging} hands this decision to the agent itself, letting it choose which segments deserve verbatim retention versus aggressive condensation.

Context-resident memory is transparent and infrastructure-free, but it carries a well-known pathology: \emph{summarization drift}.
Each compression pass silently discards low-frequency details.
After enough passes, the agent ``remembers'' a sanitized, generic version of history---precisely the kind of memory that fails on edge cases.
Extending context windows to 100k+ tokens~\citep{chen2023extending} delays the problem but cannot eliminate it, and longer contexts incur quadratic cost increases in attention.

To make this concrete: consider an agent that processes 50 user interactions per day.
After one week of rolling summarization, the raw 350-turn history has been compressed through at least three summary cycles.
A rare but critical instruction from day one---say, ``never call the production database directly''---may survive the first compression but is exactly the kind of low-frequency, high-importance detail that tends to vanish by the third pass.
The agent then proceeds to call the production database, with predictable consequences.

This is not a hypothetical failure mode; it mirrors reported issues in deployed long-running chatbots and coding assistants.
The implication is clear: for any agent expected to run for more than a handful of sessions, context-resident memory should be supplemented---not replaced, but supplemented---with an external store that preserves raw records at full fidelity.

A less obvious but equally important limitation of context-resident memory is \emph{attentional dilution}.
Even within a sufficiently large window, the LLM's attention mechanism must distribute capacity across all tokens.
As more memory content is injected, the model's ability to focus on any single piece degrades---a phenomenon empirically documented in the ``lost in the middle'' literature, where information placed in the center of a long context is recalled less reliably than information at the beginning or end.
This suggests that simply making the window bigger is not a complete solution; the agent must also \emph{curate} what enters the window, which brings us back to the fundamental need for retrieval and filtering mechanisms.

\subsection{Retrieval-augmented memory stores}
\label{sec:rag_mem}

RAG~\citep{lewis2020rag} demonstrated that pairing a generator with a non-parametric retrieval index produces strong results on knowledge-intensive tasks.
In agent settings, the store is populated not with encyclopedia articles but with \emph{living interaction records}: tool call logs, environment observations, user corrections, partial plans, and verbal reflections.

\textbf{Indexing granularity.}
Fine-grained indexing (individual tool calls or single sentences) gives precise recall but can fragment multi-step reasoning into meaningless shards.
Coarse-grained indexing (full sessions or long passages) preserves context but drowns the signal in noise.
The practical sweet spot is multi-granularity indexing, where the retriever adaptively selects the right resolution.
Dense passage retrieval~\citep{karpukhin2020dense} via learned encoders, typically backed by FAISS-style approximate nearest-neighbor search~\citep{johnson2019faiss}, remains the default implementation, often augmented with sparse BM25 and metadata filters (timestamps, tool types, task tags).

\textbf{Query formulation.}
A subtlety that many systems gloss over: the agent's immediate input $x_t$ is often a poor retrieval query.
A user asking ``Why did that crash?'' needs the agent to retrieve the crash log from two sessions ago, not the most semantically similar sentence.
Strategies include LLM-reformulated queries, multi-query fan-out with result fusion, and using the current subgoal as an additional retrieval signal.
Self-RAG~\citep{lanchantin2024selfrag} goes one step further and teaches the model to decide \emph{whether} retrieval is warranted at all---a simple gate that substantially cuts unnecessary latency.

\textbf{Scale.}
RETRO~\citep{borgeaud2022retro} and follow-up work on trillion-token datastores~\citep{raad2024scaling} suggest that retrieval memory can scale to years of interaction history without architectural changes.
The bottleneck shifts decisively from storage to \emph{relevance}: ensuring that the most \emph{useful}---not merely the most \emph{similar}---records are returned.

\textbf{Read-write memory.}
RET-LLM~\citep{sun2024reta} bridges free-form retrieval and structured storage by letting the agent write structured triplets at storage time while querying them via natural language.
This is a pragmatic compromise: schema at write time, flexibility at read time.

\subsection{Reflective and self-improving memory}
\label{sec:reflective}

Reflexion~\citep{shinn2023reflexion} introduced a deceptively simple idea: after failing a task, have the agent write a natural language post-mortem, then prepend it to the prompt on the next attempt.
No gradient updates, no reward model---just a text file of self-critiques.
The results were striking: 91\% pass@1 on HumanEval, versus 80\% for GPT-4 without reflection.

Generative Agents~\citep{park2023generative} built a richer pipeline.
Raw observations accumulate in an episodic stream.
Periodically, the agent clusters related observations and synthesizes higher-order \emph{reflections}---e.g., ``Klaus has been eating alone and seems withdrawn.''
Retrieval scores memories by a weighted mix of recency (exponential decay), relevance (embedding similarity), and importance (a self-assessed integer).
This multi-signal scoring is a substantial improvement over pure cosine similarity and remains influential in later designs.

ExpeL~\citep{zhao2024expel} pushes the paradigm further by systematically contrasting successful and failed trajectories, extracting discriminative ``rules of thumb,'' and storing them as reusable heuristics.
Think-in-Memory~\citep{zhang2024thinkinmemory} separates retrieval from reasoning: the agent first recalls, then performs a dedicated thinking step over the recalled content before generating a response.

The central risk of reflective memory is \emph{self-reinforcing error}.
If the agent incorrectly concludes ``API X always returns errors with parameter Y,'' it will avoid that call path forever, never collecting evidence to overturn the false belief.
Over-generalization is the sibling risk: a lesson learned in one context applied blindly in another.
Quality gates---confidence scores, contradiction checking against other memories, periodic expiration---are necessary but still underdeveloped.

The problem becomes more acute at scale.
A single incorrect reflection in a short-lived agent causes limited damage; the same incorrect reflection persisting in a long-running production agent---potentially influencing thousands of downstream decisions over weeks---can be catastrophic.
The severity of the reflective memory failure mode scales with agent lifetime, making it particularly dangerous in exactly the settings where memory is most needed.

One mitigation strategy explored in recent work is \emph{reflection grounding}: requiring the agent to cite specific episodic evidence for each reflection it generates.
If the reflection ``API X is unreliable'' must point to three concrete failure instances, the agent is less likely to generate baseless generalizations.
This does not fully solve the problem---the cited evidence may itself be unrepresentative---but it provides an auditable trail that can be reviewed by human operators.

\subsection{Hierarchical memory and virtual context management}
\label{sec:hierarchical}

MemGPT~\citep{packer2024memgpt} borrows an idea that operating system designers perfected decades ago: virtual memory.
An OS gives each process the illusion of vast, contiguous memory by transparently paging data between RAM and disk.
MemGPT does the same for the LLM's context window:
\begin{itemize}[leftmargin=1.5em]
    \item \textbf{Main context} (RAM): the active window holding system prompt, recent messages, and currently relevant records.
    \item \textbf{Recall storage} (disk): a searchable database of all past messages.
    \item \textbf{Archival storage} (cold storage): a vector-indexed store for documents and long-term knowledge.
\end{itemize}
The agent moves data between tiers by calling memory management ``functions''---\texttt{archival\_memory\_search}, \texttt{core\_memory\_append}, and so on.
An interrupt mechanism passes control to the agent on each user message or timer event, letting it perform multiple internal memory operations before responding.

JARVIS-1~\citep{wang2024jarvis} extends the hierarchical principle to multimodal settings, with separate stores for visual observations, textual plans, and executable skills.
Cognitive Architectures for Language Agents~\citep{sumers2024cognitive} propose a generalized blueprint where working, episodic, semantic, and procedural stores interact through a central executive (the LLM), directly echoing Baddeley's model~\citep{baddeley2000episodic}.

The Achilles' heel of hierarchical memory is \emph{orchestration}.
Page the wrong things in and you waste precious context tokens; archive too aggressively and you create ``memory blindness''---the agent simply does not know that the critical fact exists somewhere in cold storage.
This tension motivates the next mechanism family.

It is worth noting that orchestration failures in hierarchical memory tend to be \emph{silent}.
Unlike a crashed API call, which produces an error message, a paging decision that evicts the wrong record simply results in a slightly worse response---no exception, no log entry, no obvious signal that something went wrong.
Over time, these silent failures compound.
Diagnosing them requires detailed memory operation logs and retrospective analysis---an engineering investment that few current systems make but that is essential for production-grade deployments.

\subsection{Policy-learned memory management}
\label{sec:learned}

Heuristics and prompted self-control are not optimized for the agent's end task.
A $k$-nearest-neighbor retriever does not know whether the retrieved record will actually help; a fixed summarization schedule does not care whether the material being compressed is important.

Agentic Memory (AgeMem)~\citep{yu2026agentic} addresses this by treating five memory operations---store, retrieve, update, summarize, discard---as callable tools within the agent's policy, then optimizing the entire pipeline with reinforcement learning.
Training proceeds in three stages: supervised warm-up on memory demonstrations, task-level RL with outcome rewards, and finally step-level GRPO that provides denser credit assignment for individual memory actions.
Across five long-horizon benchmarks, AgeMem consistently outperforms strong baselines, and the learned policy surfaces non-obvious tactics: proactively summarizing intermediate results \emph{before} the context fills up, and selectively discarding records that are semantically similar to existing ones but add no new information.

Open concerns remain.
RL training over long horizons is expensive.
Learned forgetting could delete safety-critical information.
Policies trained on one task distribution may fail to transfer.
And it is hard to explain \emph{why} the agent chose a particular memory action---interpretability lags behind capability.

\subsection{Parametric memory and weight-based adaptation}
\label{sec:parametric}

All of the above treat memory as external to the model's weights.
An alternative family embeds memory \emph{inside} the parameters through fine-tuning or adapter modules.
MemLLM~\citep{modarressi2024memllm} fine-tunes the LLM to interact with an explicit read-write memory module, tightly coupling parametric and non-parametric knowledge.
Joint training of retrieval and generation~\citep{zhong2022training} yields better memory utilization than frozen-retriever baselines.

Parametric memory offers seamless integration---the model just ``knows'' things.
But it is hard to audit (where exactly in the weights is the user's birthday stored?), hard to delete from (machine unlearning is still immature), and expensive to update (each new fact requires fine-tuning).
For these reasons, most deployed agents favor non-parametric, inspectable stores.

\section{Evaluation: From Recall to Agentic Utility}
\label{sec:evaluation}

\subsection{Why classical retrieval metrics fall short}
\label{sec:classical}

Precision@$k$ and nDCG tell you whether the right document was retrieved.
They say nothing about whether the agent \emph{used} that document correctly---or whether retrieving it was even worth the latency.
Agent memory evaluation must jointly assess \emph{memory quality} and \emph{decision quality}, along with concerns that classical IR ignores entirely: staleness, contradiction, forgetting quality, and governance compliance.

\subsection{The new benchmark landscape}
\label{sec:benchmarks}

Four recent benchmarks push evaluation in complementary directions.

\textbf{LoCoMo}~\citep{maharana2024locomo} tests very long-term conversational memory: up to 35 sessions, 300+ turns, and 9k--16k tokens per conversation.
Three evaluation tasks---factual QA, event summarization, and dialogue generation---probe different memory demands.
The headline result: even RAG-augmented LLMs lag far behind humans, especially on temporal and causal dynamics.

\textbf{MemBench}~\citep{tan2025membench} distinguishes \emph{factual} from \emph{reflective} memory and tests each in both \emph{participation} and \emph{observation} modes.
Metrics span three dimensions: effectiveness (accuracy), efficiency (number of memory operations), and capacity (performance degradation as the memory store grows).

\textbf{MemoryAgentBench}~\citep{hu2025memoryagentbench} grounds evaluation in cognitive science, probing four competencies: accurate retrieval, test-time learning, long-range understanding, and selective forgetting.
Long-context datasets are reformatted into incremental multi-turn interactions to simulate realistic accumulation.
No current system masters all four competencies; most fail conspicuously on selective forgetting.

\textbf{MemoryArena}~\citep{he2026memoryarena} embeds memory evaluation inside complete agentic tasks---web navigation, preference-constrained planning, progressive information search, and sequential formal reasoning---where later subtasks depend on what the agent learned from earlier ones.
The most striking finding: models that score near-perfectly on LoCoMo plummet to 40--60\% in MemoryArena, exposing a deep gap between passive recall and active, decision-relevant memory use.

\subsection{Benchmark comparison}
\label{sec:bench_comparison}

Table~\ref{tab:benchmarks} summarizes design differences across these four benchmarks.

\begin{table*}[!t]
\centering
\caption{Feature comparison of recent agent memory benchmarks.}
\label{tab:benchmarks}
\small
\begin{tabular}{p{2.6cm}cccccc}
\toprule
\textbf{Benchmark} & \textbf{Year} & \textbf{Multi-session} & \textbf{Multi-turn} & \textbf{Agentic tasks} & \textbf{Forgetting} & \textbf{Multimodal} \\
\midrule
LoCoMo         & 2024 & \checkmark & \checkmark & --         & --         & \checkmark \\
MemBench       & 2025 & --         & \checkmark & --         & --         & --         \\
MemoryAgentBench & 2025 & --       & \checkmark & --         & \checkmark & --         \\
MemoryArena    & 2026 & \checkmark & \checkmark & \checkmark & --         & --         \\
\bottomrule
\end{tabular}
\end{table*}

\subsection{A practical metric stack}
\label{sec:metrics}

Deployment demands more nuance than any single benchmark provides.
We propose a four-layer evaluation stack:

\textbf{Layer 1---Task effectiveness:}
success rate, factual correctness, plan completion rate.

\textbf{Layer 2---Memory quality:}
retrieved-record precision/recall, contradiction rate, staleness distribution, coverage of task-relevant facts.

\textbf{Layer 3---Efficiency:}
latency per memory operation, prompt tokens consumed by memory content, retrieval calls per step, storage growth over time.

\textbf{Layer 4---Governance:}
privacy leakage rate, deletion compliance, access-scope violations.

Ablation studies should isolate the write policy, the retrieval strategy, and the compression module to attribute gains to specific components rather than the overall pipeline.

\subsection{Cross-cutting lessons from the benchmarks}
\label{sec:bench_findings}

Aggregating results across these four evaluations, several patterns stand out.

\textit{Long context is not memory.}
Despite context windows stretching to 200k tokens~\citep{chen2023extending}, long-context models consistently underperform purpose-built memory systems on tasks requiring selective retrieval and active management.
MemoryArena makes this starkest: passive recall aces are poor memory agents.

\textit{RAG helps, but the gap to humans is wide.}
RAG-based agents beat pure long-context baselines across the board, yet the primary bottleneck is no longer storage---it is \emph{retrieval quality}.
Agents routinely surface plausible but stale or off-topic records~\citep{maharana2024locomo}.

\textit{Nobody evaluates forgetting well.}
Only MemoryAgentBench tests selective forgetting explicitly.
Yet in any long-running deployment, the inability to discard outdated information gradually poisons retrieval precision.

\textit{Cross-session coherence is underexplored.}
Most benchmarks measure within-session performance.
MemoryArena's multi-session design reveals that maintaining consistent knowledge and behavior across sessions separated by hours or days is a distinct---and largely unsolved---challenge.

\textit{The parametric--non-parametric gap is real.}
Systems with parametric memory (fine-tuned weights) and non-parametric memory (external stores) show different failure profiles.
Parametric memory excels at seamless knowledge integration but fails at targeted deletion and auditing.
Non-parametric memory supports inspection and governance but can feel ``bolted on''---the agent sometimes ignores retrieved records or uses them inconsistently.
The optimal balance between these two approaches, and how to combine them effectively, remains an open empirical question.

\textit{Evaluation must include cost.}
A memory system that achieves 5\% higher accuracy but triples latency and storage cost may not be an improvement in practice.
None of the current benchmarks systematically report efficiency metrics alongside effectiveness, making it difficult to assess whether reported gains are ``free'' or come at significant operational expense.
Future evaluations should mandate reporting of at least token consumption and latency overhead alongside accuracy numbers.

\section{Where Memory Makes or Breaks the Agent}
\label{sec:applications}

Memory is not uniformly important.
A one-shot translation tool barely needs it; a month-long project collaborator cannot function without it.
Below we examine domains where memory is the differentiating factor.

\subsection{Personal assistants and conversational agents}
\label{sec:app_chat}

A personal assistant that forgets your dietary restrictions or re-asks your timezone every session is, at best, annoying.
MemoryBank~\citep{zhong2024memorybank} models memory decay via Ebbinghaus forgetting curves~\citep{ebbinghaus1885}: frequently accessed, high-importance memories are reinforced, while neglected ones fade.
MemGPT~\citep{packer2024memgpt} demonstrates multi-session chat with evolving user models.
The core tension in this domain is \emph{personalization without overstepping}---the agent must remember enough to be genuinely helpful without surfacing information the user considers private or forgotten.

\subsection{Software engineering agents}
\label{sec:app_code}

Coding agents assist with generation, debugging, review, and project management across codebases that may contain millions of lines~\citep{qian2024chatdev,hong2024metagpt}.
Memory requirements are steep: retain architecture decisions, track bug report histories, remember code-style preferences, and maintain a library of verified solutions.
ChatDev~\citep{qian2024chatdev} equips role-playing agents (CEO, CTO, programmer, tester) with shared memory to keep a project coherent across development phases.
MetaGPT~\citep{hong2024metagpt} structures this shared memory as standardized documents---PRDs, design specs, code modules---that persist and evolve.

The distinguishing challenge here is \emph{structural scale}: the memory system must index and retrieve relevant portions of a codebase that may span thousands of files, not just conversations.

\subsection{Open-world game agents}
\label{sec:app_game}

Minecraft and similar sandboxes are popular testbeds precisely because they demand long-horizon planning and compositional skill reuse.
Voyager~\citep{wang2023voyager} showed that an ever-growing skill library enables lifelong learning: 3.3$\times$ more unique items and 15.3$\times$ faster milestone progression than prior agents.
JARVIS-1~\citep{wang2024jarvis} extends this with multimodal memory spanning visual observations and textual plans.
Ghost in the Minecraft~\citep{zhu2023ghost} uses text-based knowledge and memory for generally capable open-world agents.

The key challenge is \emph{compositional skill reuse}: the agent must not only recall individual skills but chain them creatively to solve novel problems.

\subsection{Scientific reasoning and discovery}
\label{sec:app_science}

Scientific agents must track hypotheses, record experimental outcomes, digest literature, and revise beliefs as evidence accumulates.
Memory here acts as a hypothesis ledger and evidence accumulator.
The distinctive challenge is \emph{uncertainty-aware memory}: the agent must maintain not just facts but confidence levels, and update them correctly as new data arrives---something most current memory systems handle poorly or not at all.

\subsection{Multi-agent collaboration}
\label{sec:app_multiagent}

When multiple agents work together, memory becomes a coordination mechanism.
AutoGen~\citep{wu2023autogen} lets agents build on each other's contributions through shared context.
CAMEL~\citep{li2023camel} explores role-aware communicative agents that must remember prior agreements and collaborative history.
ProAgent~\citep{zhang2024proagent} builds proactive teammates that anticipate needs based on memory of past interactions.

Two challenges dominate: \emph{shared vs.\ private memory boundaries}---what should be visible to whom?---and \emph{consistency under concurrent writes}---what happens when two agents update shared memory simultaneously?
Current multi-agent frameworks handle shared memory in one of two ways: either all memory is shared (simple but leaks private information) or each agent maintains its own store with no cross-agent access (isolated but prevents knowledge transfer).
Neither extreme is satisfactory.
A principled middle ground would define role-based access controls over a shared memory substrate, allowing a project manager agent to see high-level summaries from a developer agent without accessing the raw code diffs.
Database-style access control lists, adapted for natural language records, are a natural but unexplored solution.

\subsection{Tool use and API orchestration}
\label{sec:app_tool}

Tool-using agents~\citep{schick2023toolformer} interact with APIs, databases, and web services.
Memory must track which tools exist, how to call them, what parameters worked last time, and which sequences of calls have been verified.
AgentBench~\citep{liu2023agentbench} evaluates agents across eight environments; agents that lose track of their command history show sharp performance drops in multi-step tasks.
DERA~\citep{nair2023dera} uses dialog-turn memory to refine tool-use strategies iteratively.

A practical hazard unique to this setting is \emph{schema drift}: when an API updates its interface, stored usage patterns become invalid.
Version tracking and schema validation on stored tool-use records are essential but rarely implemented.
In fast-moving API ecosystems, a tool-use memory system that does not handle schema drift will accumulate an increasing fraction of invalid records, progressively degrading the agent's ability to reuse past experience.

The broader point here is that tool-use memory is not just about storing ``what worked''; it is about maintaining a living, versioned catalog of tool capabilities that degrades gracefully as the external world changes.
This connects to the software engineering concept of dependency management: just as a build system must track library versions, a tool-using agent must track API versions in its memory store.

\subsection{Cross-domain memory transfer}
\label{sec:app_transfer}

An emerging direction is transferring memory \emph{across} domains---e.g., debugging heuristics learned in Python reused for Java, or time-management strategies from one user applied to another.
Tree of Thoughts~\citep{yao2024tree} provides a framework for deliberate problem-solving that could benefit from cross-domain procedural memory.
The open question is how to identify which memories generalize and which are hopelessly context-specific.

\subsection{Summary: where different memory types matter most}
\label{sec:app_summary}

The application survey reveals a clear pattern: different domains stress different memory types.
Personal assistants depend most on semantic memory (user preferences and profiles).
Software engineering agents lean heavily on procedural memory (verified code patterns and architecture decisions).
Game agents need tight integration of episodic and procedural memory (what happened $+$ what to do about it).
Scientific agents require semantic memory with explicit uncertainty tracking.
Multi-agent systems add a coordination layer that no single-agent memory design currently handles well.

No existing system provides strong support across all these profiles simultaneously, which suggests that the next leap in agent memory may come from more modular, pluggable architectures where memory components can be composed and configured per deployment rather than baked into a monolithic design.

\section{Engineering Realities}
\label{sec:systems}

\subsection{The write path}
\label{sec:writepath}

Storing every interaction verbatim is tempting and almost always wrong.
Noise---small talk, redundant confirmations, repeated greetings---degrades retrieval precision.
A well-designed write path includes:
\emph{filtering} to reject low-signal records,
\emph{canonicalization} to normalize dates, names, and quantities,
\emph{deduplication} to merge overlapping entries,
\emph{priority scoring} to rank records by task relevance and novelty,
and \emph{metadata tagging} (timestamp, source, task label, confidence) to support structured queries downstream.

The optimal filtering threshold is application-specific.
A medical agent cannot afford false negatives (missing a drug allergy mention); a casual chat assistant can tolerate them.
Between these extremes lies a spectrum: enterprise customer-support bots typically prioritize high recall for contractual commitments but accept lower recall for casual preferences, while financial advisory agents demand near-perfect recall for regulatory disclosures but can afford to forget informal chit-chat.
The write-path design should be informed by a risk analysis that maps memory failure modes to their downstream consequences in the target domain.

\subsection{The read path}
\label{sec:readpath}

Not every step needs retrieval, and not every retrieval needs the full pipeline.
Practical read-path optimizations include:
two-stage retrieval (fast BM25 or metadata filter $\rightarrow$ slower cross-encoder reranker),
retrieval-or-not gating~\citep{lanchantin2024selfrag},
token budgeting that dynamically allocates context space between memory and current task,
and cache layers for high-frequency records like user preferences.

\subsection{Staleness, contradictions, and drift}
\label{sec:staleness}

A personal assistant that sends a birthday card to a user's ex-partner at the old address is not just unhelpful---it is harmful.
Long-lived memory stores accumulate stale records, and without explicit mechanisms the agent has no way to distinguish the 2024 address from the 2022 one.

Robust systems need \emph{temporal versioning} (prefer the newest record), \emph{source attribution} (user statement $>$ agent inference), \emph{contradiction detection} (flag conflicts for resolution), and \emph{periodic consolidation} (scheduled sweeps that merge duplicates and retire stale entries).

\subsection{Latency and cost}
\label{sec:latency}

Users expect sub-second responses for simple queries.
Retrieval pipelines can easily add 200--500ms.
Common mitigations: asynchronous writes (defer storage until after the response), progressive retrieval (start generating while retrieval runs in parallel), and dynamic routing (skip retrieval for straightforward requests, engage the full pipeline only when ambiguity is high).

Xu et al.~\citep{xu2024retrieval} show that retrieving a handful of highly relevant passages into a moderate-length context often beats both pure long-context and pure retrieval approaches---a useful guideline for tuning the latency--quality tradeoff.

\subsection{Privacy, compliance, and deletion}
\label{sec:privacy}

Agent memory can harbor sensitive data: health details, financial records, private conversations.
Deployments must provide encryption at rest and in transit, per-user access scoping, automated PII redaction, configurable retention policies, and auditable deletion that removes data from every tier---including vector index entries and backup snapshots.

When memories have leaked into fine-tuned weights, external deletion is insufficient.
Machine unlearning~\citep{bourtoule2021machine,liu2024rethinking} is the only path, and it remains far from production-ready.
The intersection of agent memory governance and machine unlearning is an urgent open problem.

\subsection{Three architecture patterns}
\label{sec:arch_patterns}

In practice, agent memory systems cluster into three recurring patterns:

\textbf{Pattern A: Monolithic context.}
All memory lives inside the prompt.
Zero infrastructure, fully transparent, but capacity-capped and prone to summarization drift.
Suitable for short-lived agents or rapid prototyping.

\textbf{Pattern B: Context + retrieval store.}
Working memory in the context window; long-term records in an external vector or structured store.
A retrieval pipeline injects relevant records each step.
This is the workhorse pattern behind most production agents today: coding assistants, customer-service bots, enterprise copilots.
The engineering burden is manageable; the main challenge is retrieval quality.

\textbf{Pattern C: Tiered memory with learned control.}
Multiple tiers---context, structured DB, vector store, cold archive---managed by a learned or prompted controller.
MemGPT~\citep{packer2024memgpt} and AgeMem~\citep{yu2026agentic} are exemplars.
This pattern offers the most headroom but demands the most sophisticated engineering and training.

Our recommendation: start with Pattern~B, instrument it thoroughly, and graduate to Pattern~C only when empirical data shows that learned control meaningfully improves your target workload.

\subsection{Observability and debugging}
\label{sec:observability}

Memory systems are notoriously difficult to debug.
When an agent gives a wrong answer, was the problem in retrieval (wrong records surfaced), in the write path (relevant information never stored), in compression (detail lost during summarization), or in the LLM's reasoning over correctly retrieved content?

Production deployments benefit from comprehensive memory operation logging: every write, read, update, and delete should be recorded with timestamps, triggering context, and the records involved.
Replay tools that let developers re-run a failed interaction with modified memory content are invaluable for root-cause analysis.
Some teams have found that a simple ``memory diff''---showing what changed in the memory store between two conversation turns---provides more diagnostic value than traditional log analysis.

This observability infrastructure is rarely discussed in research papers, but its absence is one of the primary reasons that impressive demo-stage memory systems fail to make the transition to reliable production deployments.

Beyond debugging, memory observability also supports \emph{continuous improvement}.
By analyzing patterns in memory operations---which types of records are retrieved most often, which are written but never read, which retrieval queries consistently return empty results---teams can identify bottlenecks and calibrate their memory systems over time.
This feedback loop is standard in database engineering (query performance monitoring, index optimization) but is almost entirely absent in agent memory practice.
Borrowing these techniques, even in simplified form, would significantly improve the reliability of deployed memory-augmented agents.

A related concern is \emph{regression testing for memory behavior}.
When a memory system is updated---say, a new embedding model is deployed for the vector store---the effects on retrieval quality are unpredictable and potentially subtle.
Without a suite of regression tests that verify expected memory behavior on representative scenarios, changes to the memory subsystem become a source of untracked risk.
Building such test suites requires ground-truth annotations of ``which memories should be retrieved for which queries,'' an investment that pays dividends in system stability.

\section{Positioning Relative to Prior Surveys}
\label{sec:related}

Xi et al.~\citep{xi2023rise} and Wang et al.~\citep{wang2024survey} offer broad agent surveys in which memory is one module among many.
Zhang et al.~\citep{zhang2024surveymemory} focus specifically on memory and organize their review around write--manage--read operations; our work updates their coverage with 2025--2026 systems (AgeMem, MemBench, MemoryAgentBench, MemoryArena), adds a POMDP-grounded formulation, and extends the discussion to applications, engineering patterns, and governance.

Gao et al.~\citep{gao2024rag_survey} survey RAG comprehensively, but their scope is the retrieval--generation pipeline, not agent-specific memory needs.
Sumers et al.~\citep{sumers2024cognitive} propose a cognitive architecture blueprint for language agents; our taxonomy is complementary, sharing the cognitive science terminology but extending the analysis to representational substrates and control policies.

A question often raised is whether expanding context windows---from 4k in early GPT models to 200k+ today~\citep{chen2023extending}---makes external memory obsolete.
The evidence says no.
Longer context enlarges working memory but does not provide persistent cross-session storage, structured knowledge organization, selective retrieval from months of history, or governance mechanisms like deletion and access control.
Moreover, inference cost scales quadratically with context length, and Xu et al.~\citep{xu2024retrieval} show empirically that a modest context augmented with targeted retrieval outperforms brute-force long context on many tasks.

\section{Open Challenges}
\label{sec:future}

\subsection{Principled consolidation}
\label{sec:consolidation}

Current systems oscillate between hoarding (store everything, drown in noise) and amnesia (compress aggressively, lose rare but vital facts).
Neuroscience offers a suggestive model: during sleep, the hippocampus replays recent experiences, strengthening important traces and pruning the rest~\citep{squire2004memory}.
An analogous ``offline consolidation'' process---scheduled during idle periods---could provide a principled balance.
Open questions: how to estimate memory importance without future-sight, how to detect when consolidation is needed, and how to guarantee that safety-critical records survive the process.

One concrete approach worth exploring is \emph{dual-buffer consolidation}, where newly formed memories reside in a ``hot'' buffer during a probation period and are promoted to long-term storage only after passing quality checks---re-verification, deduplication, and importance scoring.
This mirrors the hippocampal-to-neocortical transfer observed in biological memory~\citep{squire2004memory}, where new memories are initially hippocampus-dependent and gradually become independent through repeated reactivation.
Implementing this in an agent system would require defining the probation period, the promotion criteria, and the fallback behavior when the hot buffer overflows before promotion occurs.

\subsection{Causally grounded retrieval}
\label{sec:causal}

Semantic similarity answers ``what looks like this?'' but not ``what caused this?''
When an agent debugs a system failure, the relevant memory may be temporally distant and semantically dissimilar to the current error message yet causally upstream.
Hybrid retrievers blending semantic similarity, temporal ordering, causal graph traversal, and counterfactual relevance remain largely unexplored.
Building them will require integrating causal discovery techniques with memory indexing---technically challenging but potentially transformative for complex reasoning tasks.

A concrete starting point would be to augment the standard vector index with a lightweight causal metadata layer.
When storing a memory record, the agent could annotate it with an estimated \emph{causal parent}---the earlier record or event that precipitated it.
At retrieval time, the system would traverse these causal links alongside the standard similarity search, surfacing records that are semantically distant but causally relevant.
Such a system need not perform full causal inference; even approximate causal annotations, generated by the LLM at write time, could substantially improve retrieval for reasoning-heavy tasks like root cause analysis, counterfactual planning, and multi-step debugging.

\subsection{Trustworthy reflection}
\label{sec:trust}

Self-reflection is a powerful adaptation mechanism that can also entrench mistakes.
If the agent falsely concludes that ``approach $A$ always fails,'' it will never test approach $A$ again---a classic confirmation bias.
Future systems need external validation (check reflections against ground truth when available), uncertainty quantification (decay confidence over time without confirming evidence), adversarial probing (periodically challenge stored beliefs with counterexamples), and expiration policies (retire unvalidated reflections after a set period).

\subsection{Learning to forget}
\label{sec:forget}

Forgetting is not a bug; it is a feature---essential for robustness, privacy, and efficiency.
Yet current systems handle it crudely: hard time-based expiration, storage-limit eviction, or nothing at all.
The research problem is to learn \emph{selective} forgetting policies that maximize long-term utility under safety and compliance constraints.
Connections to machine unlearning~\citep{bourtoule2021machine,liu2024rethinking} are critical when memories have influenced model behavior through in-context learning or fine-tuning.

\subsection{Multimodal and embodied memory}
\label{sec:multimodal}

As agents move into robotics and mixed-reality, memory must fuse text, vision, audio, proprioception, and tool state.
JARVIS-1~\citep{wang2024jarvis} provides an early example in Minecraft, but real-world embodied settings add spatial memory, real-time latency constraints, and the thorny problem of cross-modal retrieval---finding a visual memory via a textual query, or vice versa.

\subsection{Multi-agent memory governance}
\label{sec:future_multiagent}

Multi-agent systems raise questions that single-agent memory never encounters: access control over shared stores, consensus protocols for concurrent writes, and knowledge transfer mechanisms between agents with different specializations.
Current approaches rely on shared conversation logs or document stores; more sophisticated designs---distributed memory with merge semantics, hierarchical shared memory with per-agent caches---remain wide open.

\subsection{Toward memory-efficient architectures}
\label{sec:future_efficient}

Memory-augmented agents are expensive: large context windows, multiple retrieval calls per step, ever-growing stores.
Sparse retrieval (activating a tiny fraction of the store per step), compressed session vectors, memory-native architectures like Recurrent Memory Transformers~\citep{bulatov2022recurrentmem}, and retrieval-free injection via adapters~\citep{modarressi2024memllm} all point toward cheaper alternatives, though none has yet demonstrated strong agent-level performance.

\subsection{Deeper neuroscience integration}
\label{sec:future_neuro}

Current agent memory borrows cognitive science labels; deeper engagement could yield better mechanisms.
Spreading activation~\citep{anderson1983act}---where accessing one memory primes related ones---could improve retrieval beyond direct similarity.
Memory reconsolidation theory---retrieval renders a memory labile and subject to revision---could inform update mechanisms.
Ebbinghaus curves~\citep{ebbinghaus1885}, already used in MemoryBank~\citep{zhong2024memorybank}, could be extended with spaced repetition to optimize reinforcement timing.

\subsection{Foundation models for memory management}
\label{sec:future_foundation}

A longer-term vision: a \emph{foundation model for memory control}, trained across diverse agent tasks to perform write, retrieve, summarize, forget, and consolidate operations with general competence---much as instruction-tuned LLMs~\citep{ouyang2022instructgpt} provide general language capabilities.
AgeMem~\citep{yu2026agentic} takes a first step by learning memory management as a policy, but the vision of a truly task-agnostic memory controller remains unrealized.

Such a foundation model would need to handle a diverse distribution of memory challenges: short-term conversational tracking, long-term user profiling, high-frequency tool-use logging, rare but safety-critical information retention, and graceful degradation when storage budgets are exhausted.
The training data requirements are daunting---the model would need exposure to thousands of agent trajectories across dozens of domains, with ground-truth labels for memory operation quality.
Generating this training data synthetically, perhaps by having advanced LLMs retrospectively annotate which memory operations in historical traces were helpful or harmful, could bootstrap the process.

\subsection{Standardized evaluation}
\label{sec:future_eval}

The field still lacks a community-standard evaluation harness.
Each benchmark uses its own datasets, metrics, and protocols, making cross-paper comparison unreliable.
A GLUE-style shared leaderboard for agent memory---spanning conversational, agentic, and multi-session tracks, with standardized metrics from our four-layer stack---would substantially accelerate progress and reduce duplicated effort.

\section{Conclusion}
\label{sec:conclusion}

Memory has moved from a peripheral add-on to the central engineering and research challenge for LLM-based agents.
The field has traversed three generations in rapid succession: prompt-level compression, retrieval-augmented external stores, and end-to-end learned memory policies.
Evaluation has evolved in tandem---from static recall tests to multi-session agentic benchmarks that expose the gap between remembering a fact and actually using it.

This survey has offered a POMDP-grounded formalization, a three-axis taxonomy, a deep dive into five mechanism families, a structured benchmark comparison, an application-level analysis, and a practical engineering playbook.
But the hardest problems remain ahead: how to consolidate without catastrophic loss, how to retrieve by cause rather than similarity, how to reflect without entrenching errors, and how to forget safely.
Solving these will determine whether the next generation of agents is merely impressive or genuinely reliable.

If there is one takeaway from this survey, it is that memory deserves the same level of engineering investment as the LLM itself.
Model selection gets months of careful benchmarking; memory architecture often gets an afternoon.
The evidence reviewed here suggests that flipping this priority---treating memory as a first-class system component worthy of dedicated design, testing, and optimization---may be the single highest-leverage intervention available to agent builders today.

\section*{Acknowledgments}
This manuscript targets \emph{Advanced Intelligent Systems}.
Author, funding, and ethics metadata should be finalized before submission.

\section*{Conflict of Interest}
The authors declare no conflict of interest.

\section*{Data Availability Statement}
No primary data were generated in this study.
All referenced works are publicly available as cited.

\bibliography{refs}

\end{document}